\documentclass[10pt,twocolumn,letterpaper]{article}

\usepackage{cvpr}
\usepackage{times}
\usepackage{epsfig}
\usepackage{graphicx}
\usepackage{amsmath}
\usepackage{amssymb}
\usepackage[title]{appendix}
\usepackage{chngcntr}

\usepackage[pagebackref=true,breaklinks=true,letterpaper=true,colorlinks,bookmarks=false]{hyperref}

\cvprfinalcopy 

\def\cvprPaperID{6913} 

\ifcvprfinal\fi
\begin{document}


\title{Synthesizing Visual Illusions  Using  Generative Adversarial Networks}





\author{A. Gomez-Villa\thanks{Department of Information and Communication Technologies (DTIC), Universitat Pompeu Fabra. C. Roc Boronat 138, 08018, Barcelona, Spain}, A. Mart\'in\footnotemark[1],
J. Vazquez-Corral\footnotemark[1],
J. Malo\thanks{Image Proc. Lab, Universitat de Val\`encia, Val\`encia, Spain},
 M. Bertalm\'io\footnotemark[1]\\
{\tt\small \{alexander.gomez, adrian.martin, javier.vazquez, marcelo.bertalmio\}@upf.edu, jesus.malo@uv.es }
}

\maketitle

\begin{abstract}
  Visual illusions are a very useful tool for vision scientists, because they allow them to better probe the limits, thresholds and errors of the visual system. In this work we introduce the first ever framework to generate novel visual illusions with an artificial neural network (ANN).
  It takes the form of a generative adversarial network, with a generator of visual illusion candidates and two discriminator modules, one for the inducer background and another that decides whether or not the candidate is indeed an illusion.
  The generality of the model is exemplified by synthesizing illusions of different types, and validated with psychophysical experiments
  that corroborate that the outputs of our ANN are indeed visual illusions to human observers.
  Apart from synthesizing new visual illusions, which may help vision researchers, the proposed model has the potential to open new ways to study the similarities and differences between ANN and human visual perception.
  \end{abstract}


\section{Introduction}\label{sec:introduction}
A prevalent view in vision science, pioneered by the works of Attneave and Barlow in the mid-twentieth century \cite{Attneave1954,Barlow1961}, is the one given by the efficient representation principle. Stating that the organization of the visual system in general and neural responses in particular are tailored to the statistics of the images that the individual typically encounters, this principle affirms that visual information can be encoded in the most efficient way, optimizing the limited biological resources.
This is an ecological approach for vision science that has proven to be extremely successful, being the only framework able to predict the functional properties of neurons from a simple theory.
There is an agreement in considering this codification strategy as a result of an evolutionary process, a view shared by more recent and very popular theories \cite{barlow2001redundancy,friston2009free,kingdom2011lightness}.

\begin{figure}[t]
\begin{center}
\includegraphics[width=0.5\linewidth]{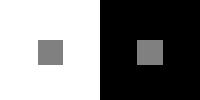}
\caption{A canonical visual illusion (brightness contrast~\cite{bruke}). The squares have the same gray value, but they are perceived as being different.}
\end{center}
\label{fig:canon}
\end{figure}

\begin{figure}[t]
\begin{center}
\includegraphics[width=1\linewidth]{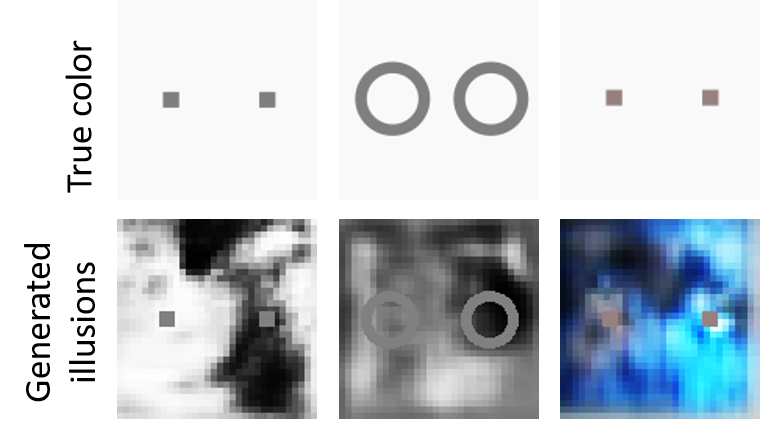}
\caption{First row targets (squares, rings, and squares) in a neutral background. Second row, visual illusions generated by our framework. Note that the values of the targets are exactly the same in the top and the bottom rows.}
\end{center}
\label{fig:illusions}
\end{figure}

The direct study of visual perception is an extremely challenging open problem, and for this reason most psychophysical research is performed on the study of perceptual limits, thresholds, and \emph{errors}, in order to shed light on the behaviour of the visual system.
A visual illusion (VI) is an image stimulus that induces a visual percept that is not consistent with the visual information that can be physically measured in the scene. An example VI can be seen in Fig. \ref{fig:canon}, that shows a canonical contrast illusion  ~\cite{bruke}: the center squares have the exact same gray value, and therefore send the same light intensity to our eyes (as a measurement with a photometer could attest), but we perceive the gray square over the white background as being darker than the gray square over the black background.
This VI, as all VIs, are images whose statistics do not correspond to the ones that are typically found in natural images, so our perception of them suffers from \emph{errors} in the (otherwise optimal) codification strategy. 
In fact, many illusions have been explained as by-products of optimal information transmission or error minimization in statistically unusual scenarios~\cite{Barlow90,Laparra15}.
Thus, VIs allow vision scientists to devise and test new vision models in their search for a better understanding of the rules that govern visual perception.

Since 2018, a handful of works 
have observed that artificial neural networks (ANN) trained in natural images can also be ``fooled'' by VIs, in the sense that their response to an image input that is a VI for a human is (qualitatively) the same as that of humans, and therefore inconsistent with the actual physical values of the light stimulus. This has been shown for VIs of very different type: motion \cite{watanabe2018illusory}, brightness and color \cite{gomez2019convolutional}, and completion \cite{kim2019neural}.

In this article we now move in the opposite direction, and propose a general framework that allows to generate novel visual illusions with an ANN, that is, the ANN will generate images that will now ``fool'' humans. In particular, we propose a framework that takes the form of a Generative Adversarial Network (GAN) \cite{Goodfellow2014}.
The generality of the model is exemplified by synthesizing VIs of different types and with different configurations of the GAN, and validated with psychophysical experiments that corroborate that the outputs of our framework are indeed visual illusions to human observers. This is a completely novel idea, to the best of our knowledge. Some examples of our results are shown in Figure \textcolor{red}{2}.

This paper puts the focus on presenting a framework for the generation of novel visual illusions, showing the capability of our model to help vision researchers. But the potential impact of our proposed approach goes well beyond that:
it could allow to study how close particular ANNs or vision models are to modelling visual perception, as we will discuss.
Our code and models will be publicly available soon.

\section{Phantasmagoria: A framework to generate visual illusions} \label{phant}
\begin{figure*}[t!]
\begin{center}
\includegraphics[width=0.85\linewidth]{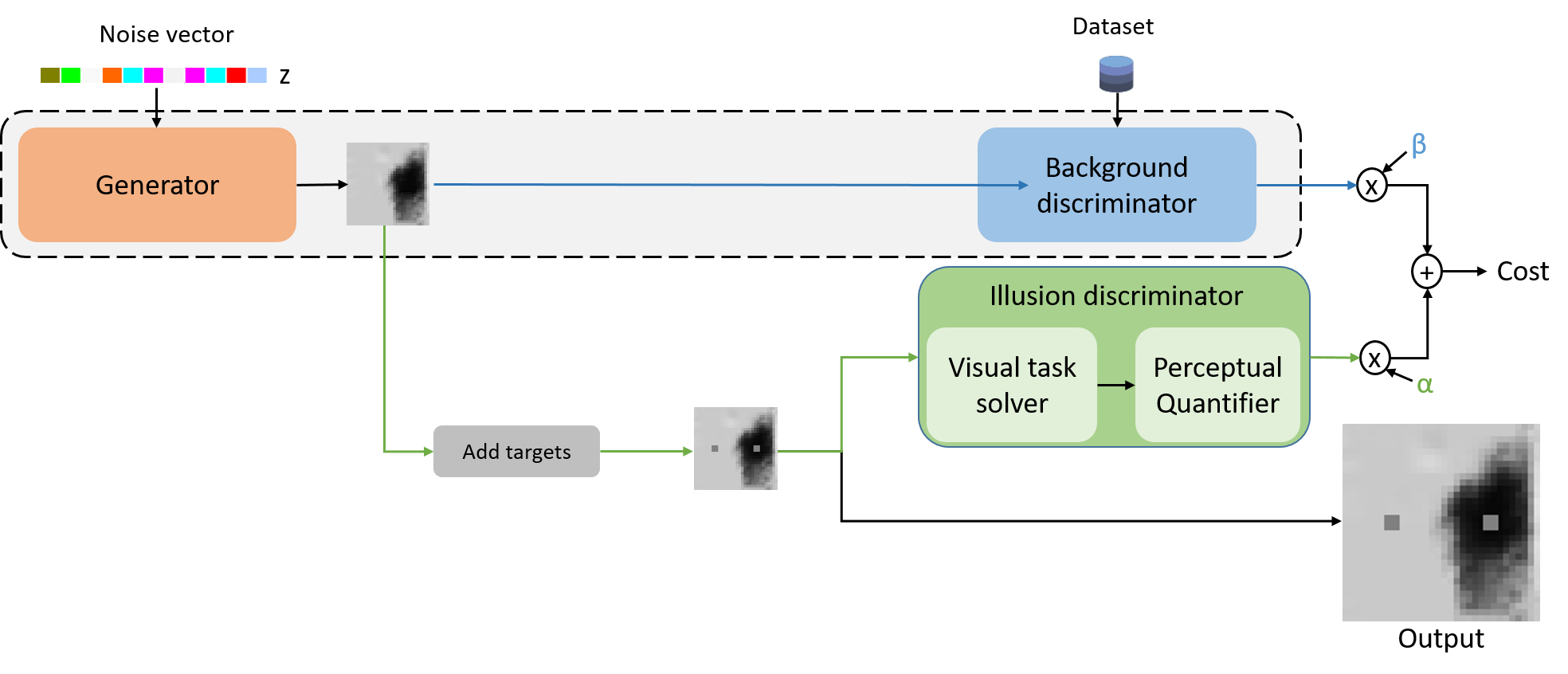}
   \caption{Phantasmagoria: A framework to generate visual illusions}
   \label{fig:framework}
\end{center}
\end{figure*}
Our goal in this work is to propose a framework that synthesize new images that produce a visual illusion to human observers. Using a generative adversarial network (GAN) approach, in principle we would just need two components: a generator of images that are candidate to produce a VI and a discriminator that ensures that it actually is a VI.
The problem with this approach is that if there is a particular candidate that produces a visual illusion with a considerably stronger effect than the rest of candidates, then the generator will end up generating many replicas of the this VI. In other words, the approach will fall into a mode collapse, thus transforming the framework into a synthesizer of the same VI again and again. This is the reason why we need a third component that pushes the generator to synthesize a variety of candidates for VIs that also follow the properties of a particular image set. 



Then the framework we propose (shown in Figure.~\ref{fig:framework}) is composed by three components, namely a candidate generator (CG), a background discriminator (BD) and an illusion discriminator (ID). The generator synthesize a candidate background (inducer) that is fed to the  background discriminator (blue branch in Figure.~\ref{fig:framework}); at the same time the targets (i.e. the region in where the illusion should happen) are pasted over the generated candidate (green branch) before being fed to the illusion discriminator. 
The background discriminator judges if the candidate agrees to a certain image type we impose. In our case, this is equivalent to deciding whether the candidate is a real instance of the training dataset - composed by textures or natural images, for example-. In contrast, the illusion discriminator plays the role of a replicant of the human vision system, in the sense that it assesses whether there is an illusion, i.e. whether it ``sees'' both targets pasted in the candidate background in a different way. The cost function for the candidate generator weightily combines the outputs from both discriminators (BD and ID) and therefore allows the CG to learn how to generate images that jointly produce a visual illusion and belong to a certain class. 

\subsection{BD: Background discriminator}

The role of the background discriminator is to force the candidate generator to synthesize inducers belonging to the type of the desired dataset. The absence of the BD module or its poor training would allow the CG to fall into collapse mode and to generate what can be considered as the ``trivial solution'' (this will be discussed in section \ref{sec:discussions}).
During the BD supervised training both images that come from the real dataset and images generated from the CG are fed into the BD with their corresponding expected probabilities. In this training scenario the BD outputs a discrepancy value. This discrepancy is computed by measuring a distance between the output probability of the input images of being from the desired dataset and the known answer of whether or not they were images from that dataset. 

\subsection{ID: Illusion discriminator}

The illusion discriminator consists of two components. The first one is a visual task solver (VTS) designed to perform a particular task that the human visual system (HVS) also achieves, and it therefore provides us with an emulation of the human visual response to the stimuli. The second component quantifies the degree of illusion present in the response given by the VTS, i.e. it acts as a perceptual quantifier (PQ). Thus, this PQ depends entirely on the type of VI to be generated and on the selected VTS for the first component.
Examples of a VTS could be for instance a CNN trained to do denoising as seen in \cite{gomez2019convolutional} or a vision model specifically designed to replicate the HVS such as the ODOG model \cite{Blakeslee1999}. 
Let us note that this ID module is the key part of our proposed framework. By selecting an ID with responses very close to those of the HVS, the framework will generate VIs that \textit{almost} always fool humans. 
Finally, this module is the part of the framework that mostly needs to be adapted in order to generate different types of illusions.  

\subsection{CG: Candidate generator}

The aim of the candidate generator is to sample images that fool both the BD and the ID. Its representation power and the sampled space are key to generate the illusion, since the inducers produced will be limited by the synthesis capability of the generator.
In our proposed framework we sample the space with the noise vector $z$, producing $n$ different inducers (which means $n$ different versions of the desired illusion). In order to have a rich variety of inducers it is recommended to pre-train the CG and BD modules (inside the dash line in the Fig.~\ref{fig:framework}) in a generative adversarial network fashion. When this pre-train is not performed, the generator risks to fall into mode  collapse. A discussion on this is presented in section \ref{sec:discussions}. 

\subsection{Framework loss function}
As discussed above, a good candidate to be a synthesized VI should balance the formation of the VI effect in the targets with the richness of the inducer background. In order to do so, our loss function for the candidate generator weights the outputs from both the ID and the BD modules using two scalars ($\alpha$ and $\beta$ respectively) as seen in Figure \ref{fig:framework}. This approach resembles what is  found in many classical models broadly used in computer vision and image processing \cite{gonzalez2002digital,forsyth2002computer}. In these classical models the loss function commonly balances two terms: one that is directly linked with the task to be solved and another one ensuring that the solution fulfills some a priori constraints, that in our case correspond to the ID and the BD modules respectively. 


\section{Generators of Visual Illusions}
In this section we detail several instances of the Phantasmagoria framework to generate visual illusions of specific types. In particular we focus on Lightness VIs (LVIs), Color VIs (CoVIs), and Contrast VIs (CrVIs). In each subsection we explain the choices for the CG, BD and ID modules required to generate each type of illusion.

In all the instances we first pre-train the CG and GD in a GAN fashion using either the DTD \cite{dtd} or the CIFAR10 \cite{cifar10} datasets. Then, both the CG and BD nets are fine-tuned with the whole framework until no significance change is observed in the loss function for any of them. Note that the ID block is never trained in this work. The $\alpha$ and $\beta$ weights are adjusted to produce high variety of illusions (as explained in section~\ref{sec:discussions}).

\subsection{Phantasmagoria-LVI: Lightness VI Generator} \label{LVI}
Lightness visual illusions (LVI) are a broadly studied topic in the vision science community. Their importance is based on its simplicity -there is only the Lightness channel- and on the importance that this Lightness channel has in our perception \cite{shapiro2016oxford}. 
In this work we focus on LVIs consisting on images that include two targets of the same shape and lightness intensity, but in which these two targets are perceived differently because of the inducer image background. 

In this instance, the CG module is a convolutional neural network (CNN) that receives a  batch (size $n$) of random noise vectors of size $100$ and generates a batch of $32\times 32$ images. The CNN is composed by two fully-connected layers followed by two convolutional layers. The fully-connected layers have $2048$ and $256\times8\times8$ hidden nodes respectively. Both convolutional layers use $5\times 5$ filter size with $128$ and $1$ channels respectively. Before each convolutional layer the input is $2 \times2$ upscaled. ReLu activation functions are used after each layer but the output one in which a sigmoid activation function is applied.
The CG is pretrained in a GAN fashion together with just the BD module. The BD module is also CNN that receives a batch of images of size $32\times 32$ and outputs their probability of belonging to the desired dataset. The network is composed by three convolutional layers and two fully-connected layers. The first convolutional layer uses $5\times 5$ filter size and the other two use $3\times 3$ filter size with $128$, $256$ and $512$ channels respectively. The two fully-connected layers have $1024$ and $1$ hidden nodes. After every layer the activation function is a Leaky ReLu ($\alpha = 0.2$) except for the output layer that uses a sigmoid function. A max pooling operation is applied after each convolutional layer. Several choices for the training dataset can be made, in sections \ref{sec:lumilu} and \ref{sec:colilu} we show the results obtained when using a database of textures \cite{dtd} or natural images like CIFAR-10 \cite{cifar10}. 

As explained in section \ref{phant}, the ID module is crucial and should be directly related with the LVI that we want to generate. In this case, for the visual task solver role we train a small CNN to perform a restoration problem (joint deblurring and denoising) inspired in the recent work of Gomez-Villa et al. \cite{gomez2019convolutional}. We call this network RestoreNet. We use the same architecture proposed there: input and output layers of size $128\times128\times3$ pixels, one hidden layer with a $5\times 5$ filter size and 8 channels, and sigmoid activation functions. The last convolutional layer works as an output layer and hence has 3 channels. We also test our framework with a different choice for the  visual task solver. In particular, we consider the ODOG model \cite{Blakeslee1999}, which is a very successful model specifically designed to replicate human vision. Since all these choices of VTS receive $128\times128$ input images the images generated by the CG are upscaled using nearest neighbor interpolation.

Regarding the perceptual quantifier part of the ID, it must measure whether there is an illusion and its strength. In this case, we opted to point-wise subtract the intensity of the central area of one target from the intensity for the same area of the other target. Please note that we can choose which target will be seen lighter by just selecting from which of the targets we subtract the other.

\subsection{Phantasmagoria-CoVI: Color VI Generator}\label{CoVI}
The color visual illusions (CoVI) that we synthesized are very related to the previously described LVI. They are images with two targets with the same shape and color that are perceived as having a different color because of the inducer image background.

The CG and BD modules are the color version of the CNNs described in the previous section, i.e. the only changes are that the output layer for the CG network has 3 layers and that the size of the inputs of the BD is of $32\times32\times3$.
For the visual task solver of the ID module we choose a recent denoising CNN proposed by Zhang \emph{et al.} \cite{Zhang2017} that according to the work of \cite{gomez2019convolutional} obtained a good replication for color illusions. The perceptual quantifier used here is a  point and channel-wise subtraction of the image values of the central area of one target from those of the same area of the other target. In order to choose a desired perceived color each channel must be tailored manually. For instance, if the desired output is a reddish right target (with respect to the left target) a possible perceptual quantifier will be one that maximizes the difference between the targets in the red channels (right minus left) and leave the blue and red channels as a free choice for the model.

Note that color VIs are highly complex, and there is still ongoing discussions on how to categorize them~\cite{otazu2010toward}. Hence, if the target is not neutral gray the framework may suggest solutions that saturate the color of one target or that only change the luminance. To alleviate this, we ask the framework to synthesize a simple perception of color, that is, show one target of one color. Therefore, mixed classes (saturation contrast or color contrast) of color illusions are expected.

\subsection{Phantasmagoria-CrVI: Contrast VI Generator}\label{CrVI}
Contrast illusions are different in nature to lightness and color ones. They are given by the fact that in humans the presence of certain patterns in the scene (previous in time, or close in space) leads to changes in the response of texture sensors \cite{ross1991contrast,foley1997analysis} and motion sensors \cite{morgan2006predicting}. For instance, in the texture aftereffect \cite{blakemore1969adaptation,barlow1990theory} parts of stimuli with physically stationary contrast seem to fade out after prolonged exposure to localized high contrast patterns of similar frequency and orientation. In the case of texture \emph{induction}, patterns with similar frequencies and orientation strongly reduce the response of sensor tuned to similar patterns \cite{cavanaugh2002nature}. 

In this work we focus on the case of texture \emph{induction}. In particular, we focus on CrVIs consisting on images that include two targets that present the same spatial frequency and orientation, but in which these two targets are perceived differently because of the inducer image background.

In this instance the CG and BD modules are the same CNNs described in the section~\ref{LVI}. Regarding the ID, we use RestoreNet as VTS. For the perceptual quantifier, we measure the Michelson's  contrast~\cite{michelson1995studies} in  the central area of both targets and maximize the difference between them. As a result we ask the framework to generate an inducer that remarks or makes more ``visible'' (i.e. with higher contrast) one of the patterns with respect to the other.

\section{Qualitative Results}\label{sec:results}
Luminance and Color VIs can be further subdivided into assimilation and contrast ones. In assimilation the target intensity gets attracted by the inducer (e.g. a gray target gets brighter if it is surrounded by a white inducer). Contrast is the opposite effect of assimilation, therefore in this case, the target gets expelled from the inducer (e.g. a gray target gets darker if it is surrounded by a white inducer). In order to include the most variate set of examples, we show how our framework performs both of these sub-types for the luminance case. In the case of color we focus on showing how our framework generate color contrast illusions independently from the starting color of the given targets. Finally, we also show how our framework generates contrast illusions -in particular texture induction- for targets of different orientations. 

We recommend the reader to look at all the different visual illusions in isolation -by for example covering any other illusion located close to the one of interest-. Also, let us note that all the visual illusions are presented in isolation in the supplementary material, together with a larger set of results.

\subsection{Lightness illusions}\label{sec:lumilu}
 Our results were obtained by selecting RestoreNet as the visual task solver, and considering different training datasets for the background discriminator -DTD \cite{dtd} and CIFAR10 \cite{cifar10}-.
 
\begin{figure}[t!]
    \centering
        \includegraphics[width=0.49\textwidth]{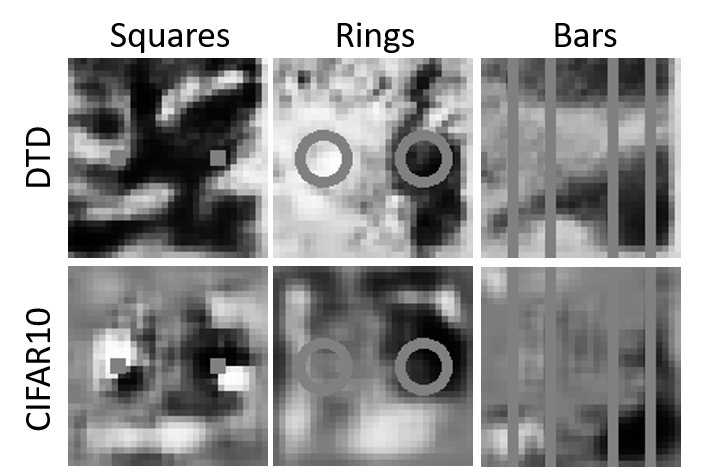}
    \caption{Results for the Lightness contrast illusions using different training datasets and shapes, and RestoreNet as the VTs. In all the cases the right shape was selected to be brighter than the right one.}
    \label{Fig:Gray_results_color}
\end{figure}

The results for the lightness contrast illusions are shown in Figure \ref{Fig:Gray_results_color}. Three different shapes for the targets -squares, rings and bars- were selected. Also, for these images, we selected that the right target should appear brighter that the left one. We can see how our framework is able to generate contrast VIs under all these different setups. Regarding the shapes, our method presents a slightly worse performance for the case of the bars, due to their large height. In terms of the training dataset, we can see how this affects the background. In particular, we can see that a more structured pattern appears in the background image inducer when the BD is trained using CIFAR10. This is probably caused by the object categories presented in this dataset.

\begin{figure}[t!]
    \centering
        \includegraphics[width=0.25\textwidth]{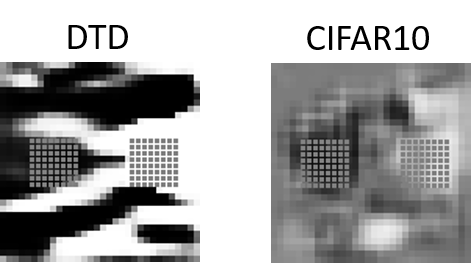}
    \caption{Results for the Lightness assimilation illusion using different training datasets, and RestoreNet as the VTS.}
    \label{Fig:Gray_results_assimilation}
\end{figure}

Figure \ref{Fig:Gray_results_assimilation} shows the result of our framework for a lightness assimilation illusion. Please note that assimilation VIs extremely depend on spatial frequency, and thus they depend on the distance at which they are viewed. They may be converted into contrast VIs if the spatial frequency -or equivalently the viewing distance- is diminished \cite{helson63,fach86}. We have done our best to accommodate the size of this figure to account for the best experience of the visual illusion. This said, in case the reader does not perceive the illusion -and as the perception of visual illusions varies between observers- we recommend him/her to zoom out the Figure to better accommodate the frequency of this illusion.

\begin{figure}[t!]
    \centering
        \includegraphics[width=0.49\textwidth]{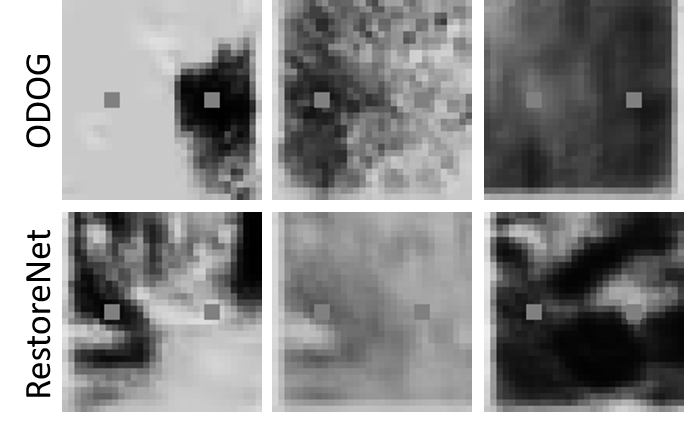}
    \caption{Results for the Lightness inductions using the DTD dataset and squares for two different VIs. These images are also a subset of those used in the psychophysical experiment. For this reason, the square selected as the one being brightest was randomized in these images}
    \label{Fig:Example_IDS}
\end{figure}

Finally, figure \ref{Fig:Example_IDS} shows the result of our framework for the two different VTS choices explained in section \ref{LVI}. In this case, the training dataset was DTD and the shape selected was square, therefore aiming for a lightness contrast illusion. Please note that in these images the square selected as the brightest was randomized in order to use these images in the psychophysical experiment explained in section \ref{psyco}. This said, we can clearly see that both of the VTS are able to obtain different VIs -although not all of them with the same strength-.

\subsection{Color illusions}\label{sec:colilu}
\begin{figure}[t!]
    \centering
        \includegraphics[width=0.49\textwidth]{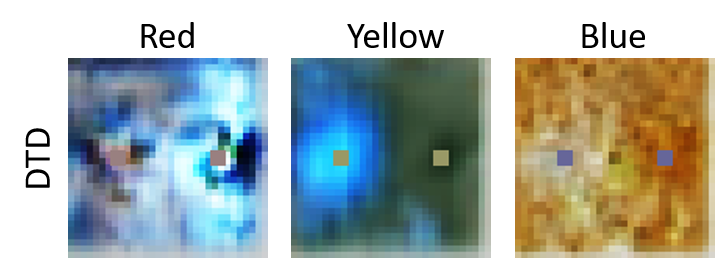}
    \caption{Results for the color illusions using DTD as the training dataset for the BG and different target colors. From left to right, our cost imposed the perception of the right target to be: redder, yellower, bluer. }
    \label{Fig:Color_results}
\end{figure}

 As we explained in section \ref{CoVI}, in this case we have selected the denoising CNN proposed in Zhang \emph{et al.} \cite{Zhang2017} as the VTS. We have considered the DTD as the training dataset for the background discriminator. The shape target selected are squares. Results are shown in Figure \ref{Fig:Color_results}. The three columns of this Figure show the color contrast illusion obtained by starting either in a blue, yellow or red target color, and thus this Figure shows that our framework is not tied to any particular region of the color space. 
 
More in detail, the three cases perform as it was imposed in the loss function. In the first column the right target is perceived redder than the left one. Similarly, in the second column the right target is perceived yellower, and in the third column the right target is perceived bluer.

\subsection{Contrast illusions}
\begin{figure}[t!]
    \centering
        \includegraphics[width=0.49\textwidth]{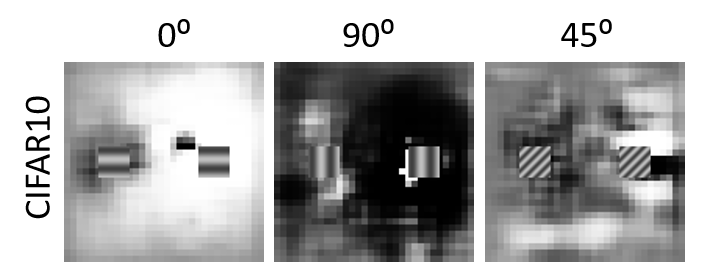}
    \caption{Results for the contrast illusions using CIFAR10 as the training dataset for the BG and different orientation patterns.}
    \label{Fig:Contrast_results}
\end{figure}

For the contrast illusions we have selected the CIFAR10 dataset as the training one for the background discriminator. We have selected CIFAR10 because the DTD dataset does not present enough high frequency training data to synthetize these type of illusions. For the visual task solver we have selected RestoreNET. In terms of the targets, we have selected patterns of orientation $0^{\circ}$, $45^{\circ}$, and $90^{\circ}$. For these images, it was selected that the right target should look higher contrasted than the left one. Results for these illusions are shown in Figure \ref{Fig:Contrast_results}. We can see that for the $0^{\circ}$ and $90^{\circ}$ cases the visual illusions accomplish our condition (i.e. the right target is higher contrasted than the left one). However, in the case of $45^{\circ}$ even if the framework performs as expected (i.e. selecting a higher frequency for surrounding the pattern of the left and a lower frequency for surrounding the pattern on the right) the effect is not strong enough to provoke an illusion for a human observer.

\section{Psychophysical tests}\label{psyco}
We have run a phychophysical test to evaluate the ability of our framework to fool human observers. Observers were shown the lightness visual illusions created by Phantasmagoria-LVI -using squares as targets-, and they were asked to select the lighter square, having three different options to choose from: left, right, or center (in case they were not able to perceive any difference). The experiment was performed on a calibrated AOC I2781FH LCD monitor. Observers were sit at 50 cm of the screen so as the target grey squares surrounded 1.5 degrees of visual angle. We have considered the Describable Textures Dataset (DTD) \cite{dtd}  for training the background discriminator module. For the visual task solver in the illusion discriminator module we have considered the two choices mentioned in section \ref{LVI}: RestoreNet \cite{gomez2019convolutional}, and the ODOG model \cite{Blakeslee1999}. For each VTS choice we have selected $50$ output images (totalling 100 images) by randomly selecting images that were considered to be an illusion by the perception quantifier module from batches of different iterations during the framework training, in order to ensure the diversity of the candidates generated. These images were randomized both in terms of the methods and in terms of the side in which the lighter square was expected to appear. Six example images (3 for each of the IDs tried: first row ODOG, second row RestoreNet) are shown in Figure \ref{Fig:Example_IDS}. The full set of images used in this experiment are shown in the supplementary material.

\begin{figure*}[t!]
    \centering
    \begin{tabular}{ccc}
    \includegraphics[width=0.3\textwidth]{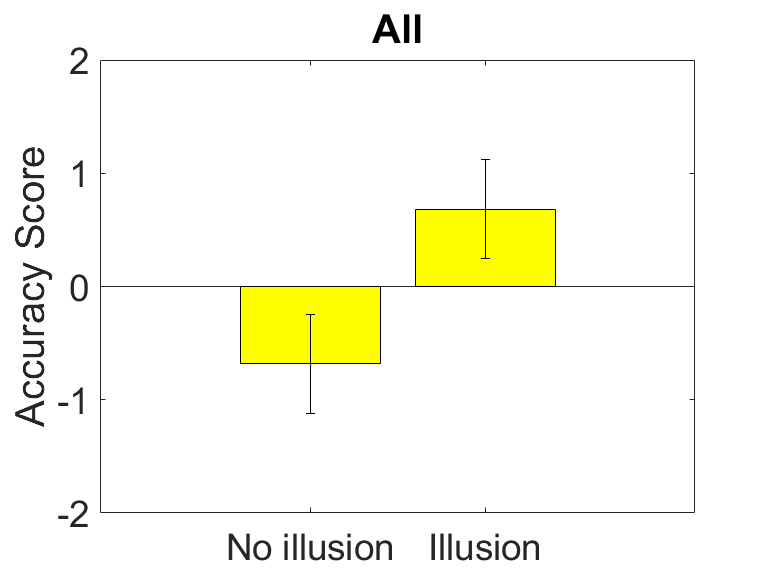} &
    \includegraphics[width=0.3\textwidth]{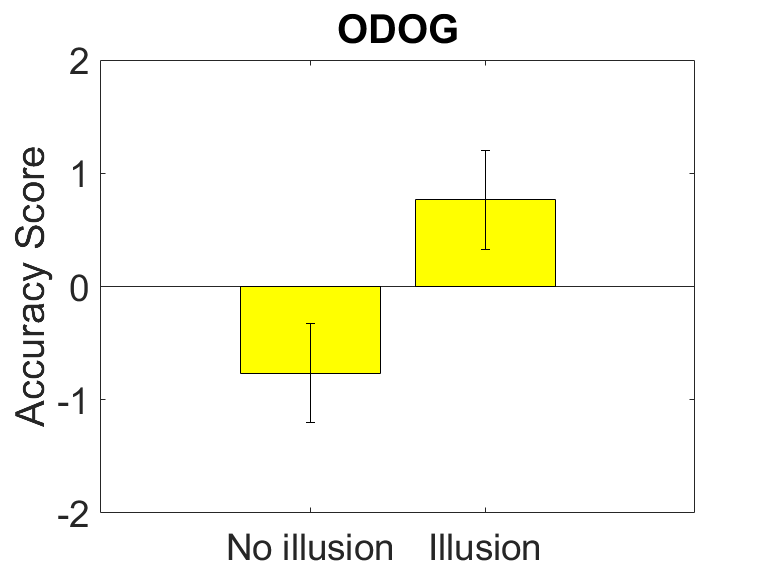} &         \includegraphics[width=0.3\textwidth]{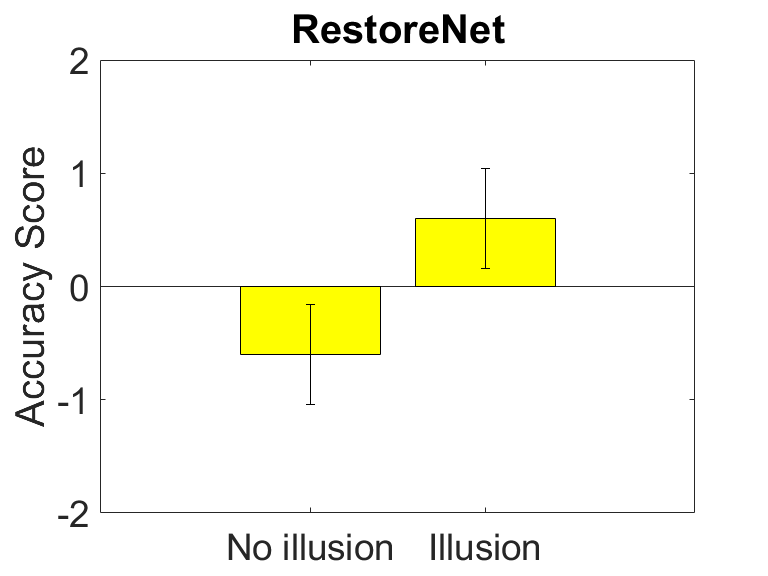}
    \end{tabular}
    \caption{Thurstone Case V Results for our Experiment. We can see that the illusions are saw with statistical significance in all the cases.}
    \label{Fig:Thurstone}
\end{figure*}

Ten observers took part in the experiment. All of them presented normal or corrected to normal color vision. None of them was an author in this paper. A first interpretation of the results is presented in Table \ref{Table:ExpPsy1}, where the average selection of each option for the full set, as well as for each ID, is shown. As we can see, in a large majority of the cases ($>66\%$) the human observers perceive the illusion that was generated by the ID. To obtain a more statistically significant result, we have also recast the experiment in terms of the Thurstone Case V Law of Comparative Judgement \cite{thurstone1927law}. To this end, we have divided the answers as having the generated illusion, or not having it. This second category considers both the cases where an observer has not seen any illusion and where an observer has selected the opposite direction for the illusion. Results for this analysis are shown in Figure \ref{Fig:Thurstone}. As we can see, the generated illusion is perceived with statistical significance in all the cases.

\begin{table}[t!]
\centering
\begin{tabular}{|l|l|l|l|}
\hline
           & \begin{tabular}[c]{@{}l@{}}Opposite\\ illusion\end{tabular} & \begin{tabular}[c]{@{}l@{}}No\\ illusion\end{tabular} & \begin{tabular}[c]{@{}l@{}}Correct\\ illusion\end{tabular} \\ \hline
All        & 0.0770  & 0.2280 & \textbf{0.6950} \\ \hline \hline
ODOG       & 0.0720  & 0.2160 & \textbf{0.7120}  \\ \hline
RestoreNet      & 0.0820  & 0.2400 & \textbf{0.6780} \\ \hline
\end{tabular}\label{Table:ExpPsy1}\caption{Results of the psychophysical experiment as average of the selected square.}
\end{table}

\section{Discussion and limitations}\label{sec:discussions}
\begin{figure}[t!]
    \centering
        \includegraphics[width=0.49\textwidth]{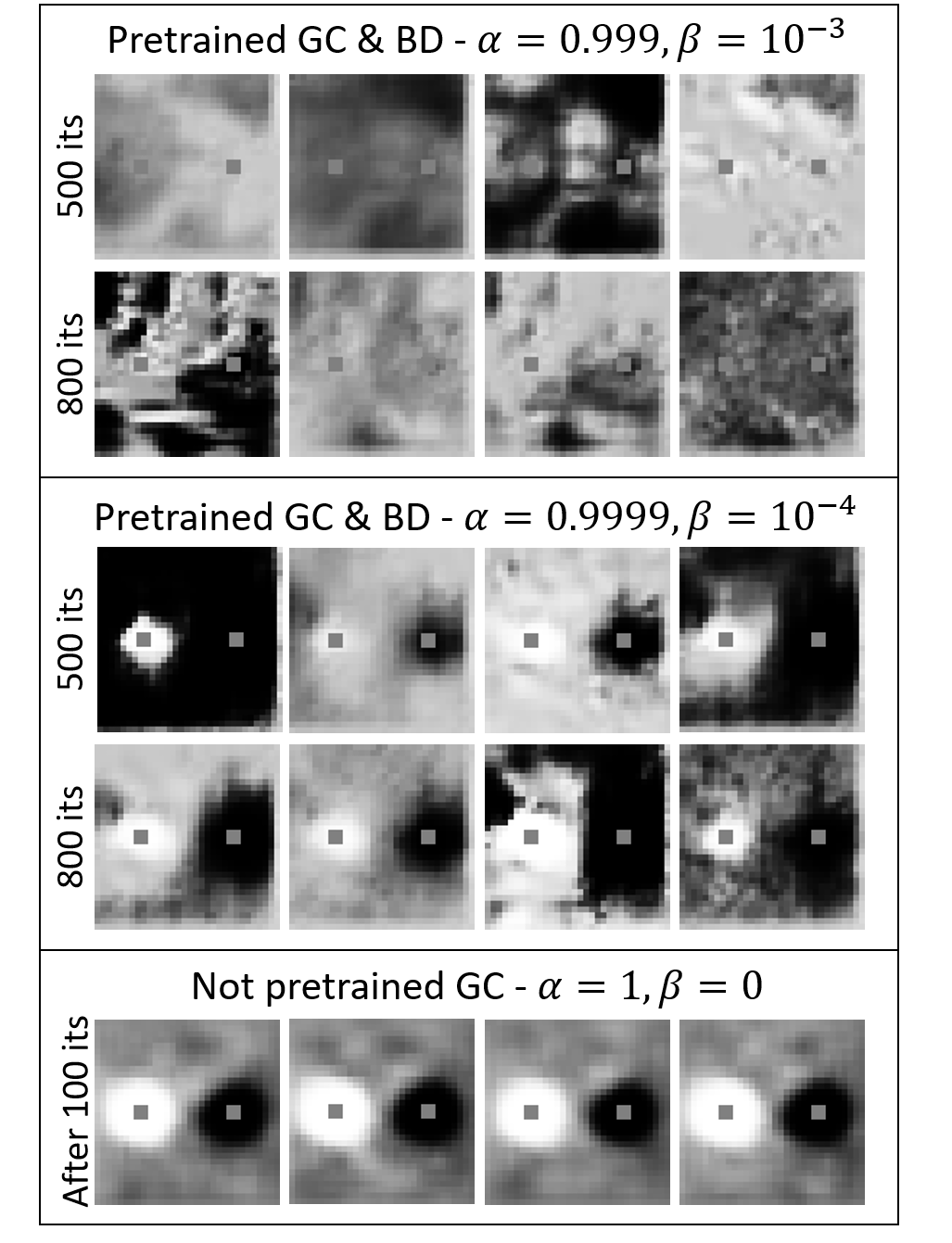}
    \caption{Parametric study of the Phantasmagoria-LVI framework using RestoreNET as Visual Task Solver.}
    \label{Fig:collapse}
\end{figure}

One of the key points of the proposed framework, as mentioned in Section \ref{phant}, is the right balance between the two discriminator modules, the background discriminator (BD) and the illusion discriminator (ID). We have studied the behavior of the proposed framework for a specific instance of Phantasmagoria-LVI in which we have chosen RestoreNET as the visual task solver and the DTD dataset \cite{dtd} for the BD. Finding a right balance between the weights of the ID ($\alpha$) and the BD ($\beta$) modules produces a rich variety of candidates to be a VI (see the top group of images in Figure \ref{Fig:collapse}). There we can see some randomly selected images from the generated images after 500 and 800 iterations of the learning process. If the weight of the BD module is too low (as shown in the middle group of images in Figure \ref{Fig:collapse}), the generated candidates will be very similar between them since these solutions will be mainly governed by the ID module. These candidates already resemble strongly the previously referred as \textit{canonical solution} for lightness VI (see Figure \ref{fig:canon}). This ``convergence'' to this \textit{canonical solution} is even more clear if we observe the extreme case of our framework: when we ignore the BD module (i.e. we set its corresponding weight to be 0) and we use a generator that has not been pretrained to generate images of a certain type. In this extreme case, in a few iterations the network falls into a mode collapse generating all the images of every batch exactly equal (as shown in the bottom group of images in Figure \ref{Fig:collapse}). The images produced by the Phantasmagoria-LVI framework in mode collapse are nothing but a version of the \textit{canonical solution} that produces a very strong illusion in human observers.

The study above introduces one general limitation of the proposed framework: there is no general rule to choose what is the best balance between the two discriminator modules. This adds up to a shared problem of most of the GAN approaches that is that of the selection of parameters such as the learning rates of the generator and discriminator. In total, this implies that obtaining good solutions requires a certain degree of fine tuning of the model parameters.

The previous discussion has also implications in the way the data for the psychophysical experiment was generated. Since the aim of our work is to synthesize new VI and hence avoid falling in this mode collapse (that we know that will produce a VI to human observers), we have to tune the framework parameters to obtain a good trade-off between a sufficiently rich variety of candidates and the strength of the effect that they provoke. The images used for the experiment are a consequence of this compromise and were randomly picked between the generated candidates instead of being manually chosen to maximize the effect in the observers. The fact that still under these conditions the observers perceived the expected VI in roughly a 70\% of the cases is a very remarkable finding.
Another by-product of the need of manually tuning the parameters in order to find this balance is that we can not use the results from the psychophysical tests to directly compare the two instances of the VTS used (RestoreNET and ODOG) in terms of how close are they to be a good ``replicant''. We have no way to ensure that the selected parameters $\alpha$ and $\beta$ are \textit{optimal} for each method in the sense that they are the ones that would provoke stronger VI in the human observers. Let us note though, that obtaining a rule to fix these parameters would open the door to cast our framework as a proxy to assess how close a given model is to modelling the human visual perception, therefore becoming a really useful tool for vision researchers.

Besides the above mentioned topics there are other important open problems that mainly concern the illusion discriminator (ID) and are intimately related with well-known open problems for the vision science community. The first one is finding a vision model that correctly replicates the human visual perception in most scenarios. This is directly related with the visual task solver (VTS) part of our ID module, which in the absence of this ``perfect model'' we substitute it by CNNs or one of the existing vision models. The second one is our need for quantitatively evaluate the strength of the illusion after they pass through the VTS. There is a well known lack of quantitative measurements in the study of VIs in the vision science literature, in where qualitative assessments are typically performed. This forces us to propose ways to measure these effects in this work and therefore implies the need for further study of the impact of the choices made in the generation of VIs.

Finally, let us remark that the possibility of using a CNN as the VTS part of the illusion discriminator opens a possibility still unexplored of allowing the VTS to be trained during the framework optimization process. This raises further questions regarding what could be the best training for improving the generation of visual illusions. 

\section{Conclusions}\label{sec:conclusions}
  We have introduced Phantasmagoria, the first ever framework to generate novel visual illusions using  artificial neural networks. Our framework follows a GAN structure, that in particular has a generator of inducer candidates to generate a visual illusion and two discriminator modules, one that ensures that the candidate belongs to a desired type of images and another one that decides whether or not the candidate provokes indeed an illusion.
  We have shown the generality of our framework by  synthesizing illusions of different shapes and types, namely lightness, color and contrast visual illusions. Furthermore, we have corroborated the validity of our approach with psychophysical experiments that confirm that most of the visual illusions produced by our framework fool in the same way human observers. 
  
  Further work should start by a deeper study of the parameters governing the model that could open the door to exciting new ways of research connecting human visual perception and artificial neural networks. 
  The extension of this work to other types of illusions should be considered. For example, we can consider the works of Watanabe \emph{et al.} \cite{watanabe2018illusory} and Kim \emph{et al.} \cite{kim2019neural}, where they found CNNs replicating motion or completion illusions respectively. Both of these CNNs seem to be good candidates for our visual task solver module. 

\section*{Acknowledgements}

This work has received funding from the European Union’s Horizon 2020 research and innovation programme under grant agreement number 761544 (project HDR4EU) and under grant agreement number 780470 (project SAUCE), and by the Spanish government and FEDER Fund, grant ref. PGC2018-099651-B-I00 (MCIU/AEI/FEDER, UE). We gratefully acknowledge the support of NVIDIA Corporation with the donation of the Titan Xp GPU used for this research.

\clearpage

{\small
\bibliographystyle{ieee}
\bibliography{egbib}
}

\clearpage
\counterwithin{figure}{section}
\begin{appendices}

\section{CNN architectures from Phantasmagoria-LVI/CoVI/CrVI}
In this section we depict the CNN architectures considered in the paper for the specific instance of the Phantasmagoria framework to synthesize lightness, color and contrast visual illusions (Phantasmagoria-LVI/CoVI/CrVI respectively). 
Fig.~~\textcolor{red}{A1} shows the candidate generator network. Fig.~\textcolor{red}{A2} depicts background discriminator network. Finally, in Fig.~\textcolor{red}{A3} we detail the visual task solver (which is part of the illusion discriminator module) and that we denote as RestoreNET.

\begin{figure*}[t!]
\begin{center}
\includegraphics[width=0.8\linewidth]{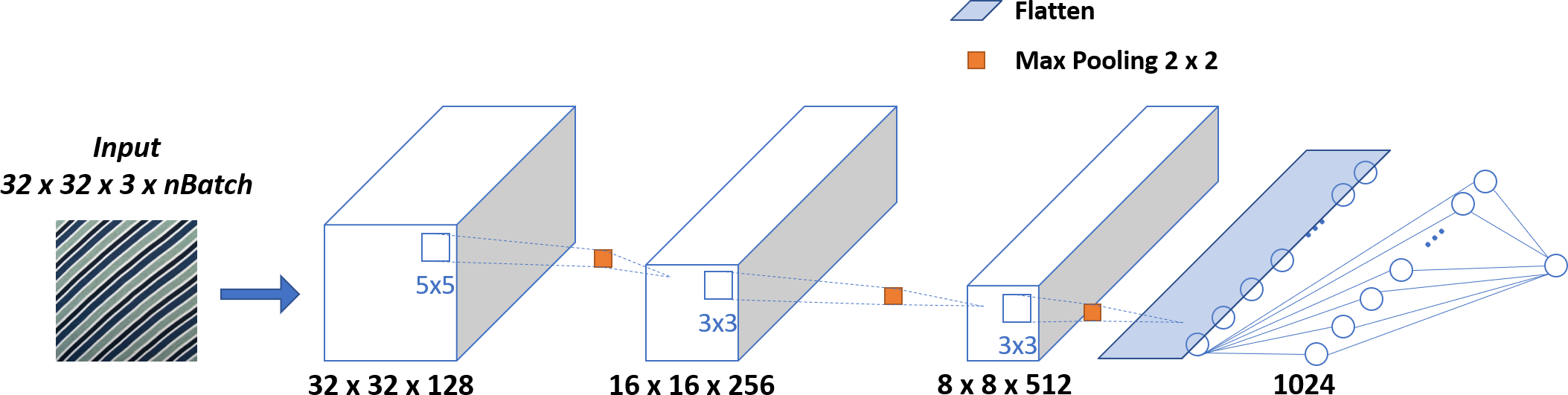}
\caption{Architecture of the background discriminator CNN}
\end{center}
\label{fig:disc}
\end{figure*}

\begin{figure*}[t]
\begin{center}
\includegraphics[width=0.8\linewidth]{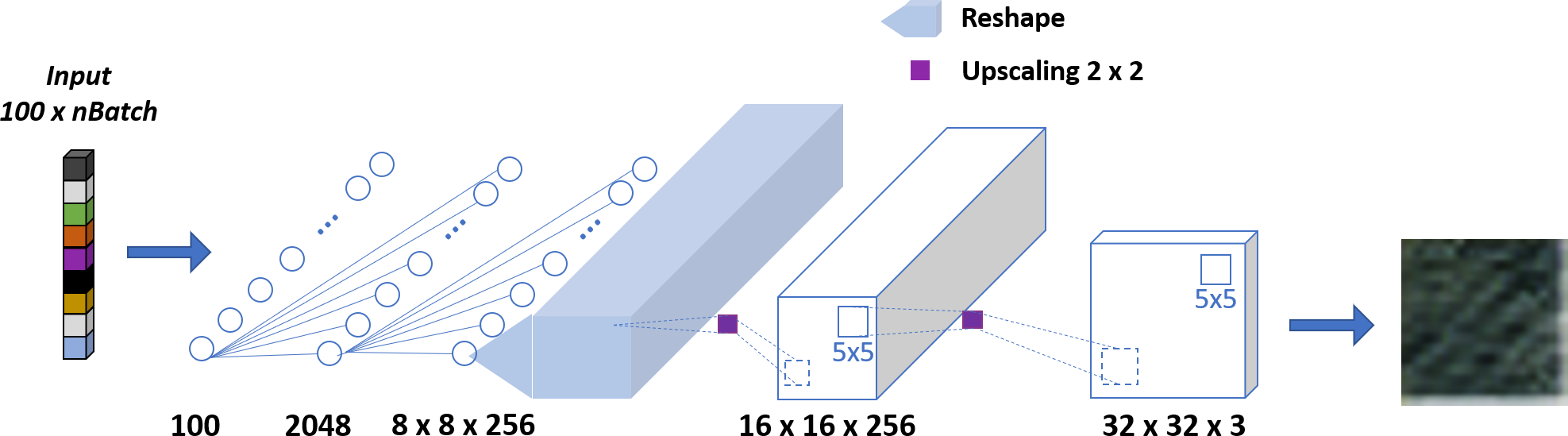}
\caption{Architecture of the candidate generator CNN}
\end{center}
\label{fig:gen}
\end{figure*}

\begin{figure*}[t]
\begin{center}
\includegraphics[width=0.8\linewidth]{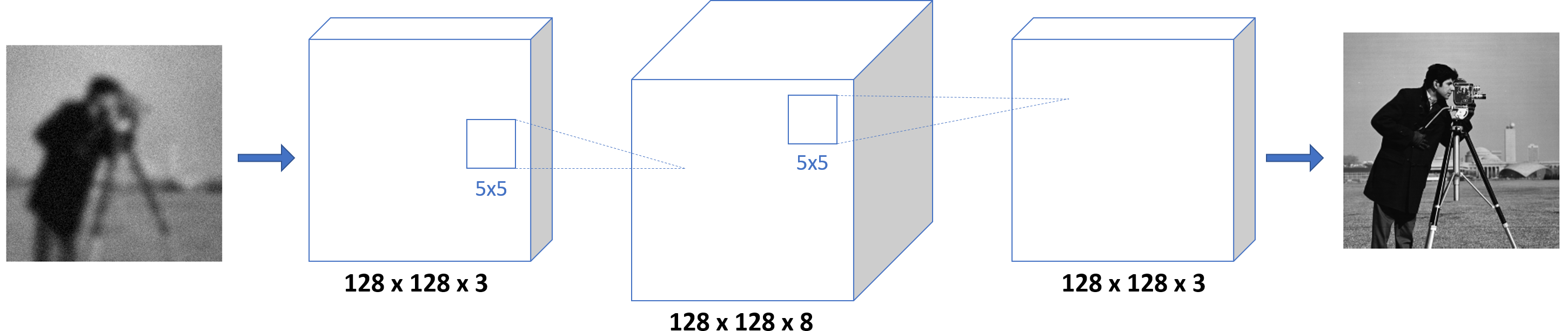}
\caption{Architecture of the visual task solver (RestoreNet) CNN}
\end{center}
\label{fig:rest}
\end{figure*}

\section{Implementation details}

All our CNN were trained in Keras~\cite{chollet2015keras} and Tensorflow~\cite{tensorflow2015-whitepaper} frameworks. We used the following loss functions: binary cross-entropy for the background discriminator network and mean squared error for the candidate generation and visual task solver. The maximum number of epochs was set to 100, and we set a batch size of 32. The training stop criteria of the framework varies depending of the value of $\alpha$ and $\beta$ parameters (please go to section 6 of the paper for more details).

The oriented difference of Gaussians model (ODOG) used was a Tensorflow implementation of an public available~\cite{odogImp} Python implementation

Our code and models will be publicly available soon.

\section{Psychophysical test images}

Figures \textcolor{red}{C1} and \textcolor{red}{C2} show the full set of images ($50$ per method) that were used in the psychophysical test (see section 5 of the paper). The square selected as the one being brightest was randomized in these images for experimental purposes.

\begin{figure*}[t]
\begin{center}
\includegraphics[width=0.6\linewidth]{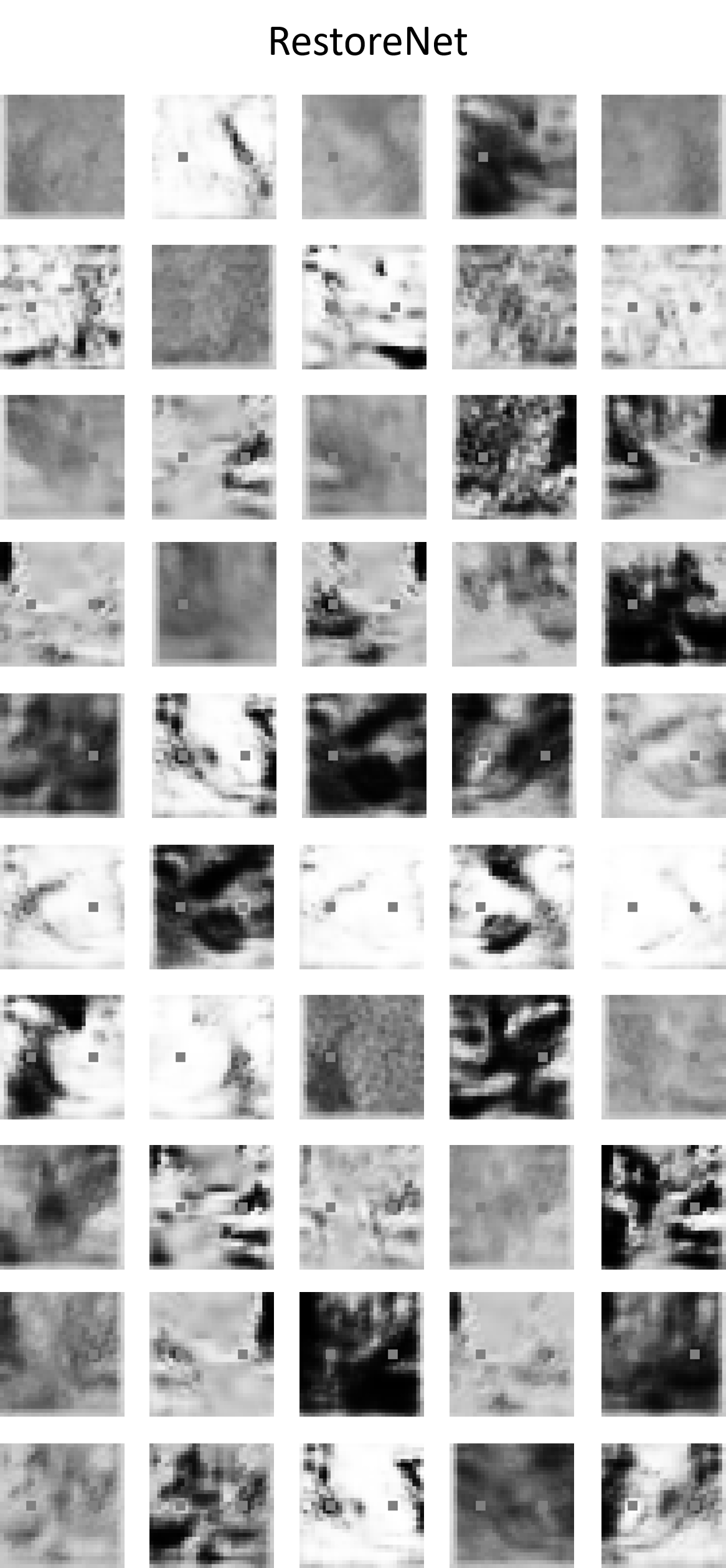}
\caption{Visual illusions generated using RestoreNet as visual task solver.  The square selected as the one being brightest are randomized in these images for experimental purposes.}
\end{center}
\label{fig:restP}
\end{figure*}

\begin{figure*}[t]
\begin{center}
\includegraphics[width=0.6\linewidth]{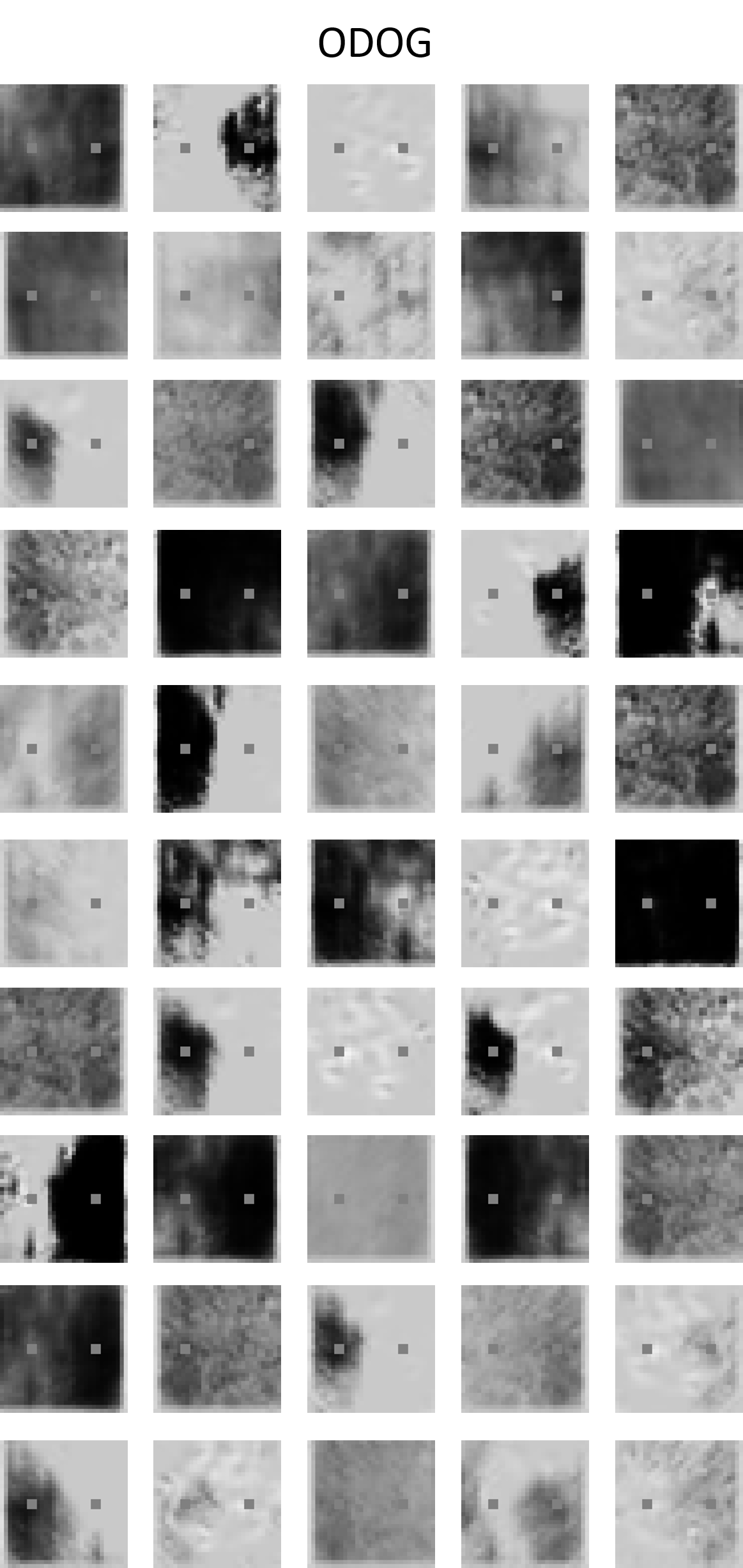}
\caption{Visual illusions generated using ODOG as visual task solver.  The square selected as the one being brightest are randomized in these images for experimental purposes.}
\end{center}
\label{fig:ODOGP}
\end{figure*}

\section{Visual illusions  in isolation and additional results}

Figures \textcolor{red}{D1} to \textcolor{red}{D14} show the different images presented in the paper in isolation. This allows the reader to better accomodate for the effects provoked by the visual illusions.  Additionally, Fig.\textcolor{red}{D15} to Fig.\textcolor{red}{D23} present several new visual illusions (different from the main paper results). The expected visual illusion is explained at each figure caption. 

Please remember that -as is always the case in visual illusions- both targets have exactly the same pixel values.


\begin{figure*}[ht]
\begin{center}
\includegraphics[width=0.15\linewidth]{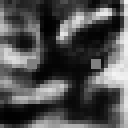}
\caption{\textbf{Right target is asked to be lighter.} LVI generated using RestoreNet as VTS and the DTD as database for the BD.}

\vspace{5cm}

\includegraphics[width=0.15\linewidth]{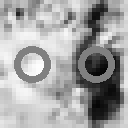}
\caption{\textbf{Right target is asked to be lighter.} LVI generated using RestoreNet as VTS and the DTD as database for the BD.}

\vspace{5cm}

\includegraphics[width=0.15\linewidth]{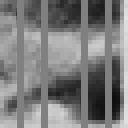}
\caption{\textbf{Right target is asked to be lighter.} LVI generated using RestoreNet as VTS and the DTD as database for the BD.}

\end{center}
\label{fig:rest}
\end{figure*}

\begin{figure*}[ht]
\begin{center}
\includegraphics[width=0.15\linewidth]{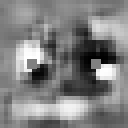}
\caption{\textbf{Right target is asked to be lighter.} LVI generated using RestoreNet as VTS and CIFAR10 as database for the BD.}

\vspace{5cm}

\includegraphics[width=0.15\linewidth]{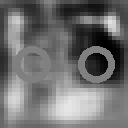}
\caption{\textbf{Right target is asked to be lighter.} LVI generated using RestoreNet as VTS and CIFAR10 as database for the BD.}

\vspace{5cm}

\includegraphics[width=0.15\linewidth]{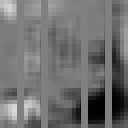}
\caption{\textbf{Right target is asked to be lighter.} LVI generated using RestoreNet as VTS and CIFAR10 as database for the BD.}

\end{center}
\label{fig:rest}
\end{figure*}

\begin{figure*}[ht]
\begin{center}
\includegraphics[width=0.15\linewidth]{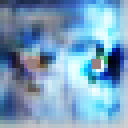}
\caption{\textbf{Right target is asked to be reddish.} CoVI generated using RestoreNet as VTS and the DTD as database for the BD.}

\vspace{5cm}

\includegraphics[width=0.15\linewidth]{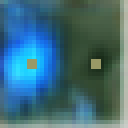}
\caption{\textbf{Right target is asked to be yellower.} CoVI generated using RestoreNet as VTS and the DTD as database for the BD.}

\vspace{5cm}

\includegraphics[width=0.15\linewidth]{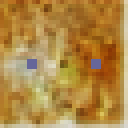}
\caption{\textbf{Right target is asked to be bluer.} CoVI generated using RestoreNet as VTS and the DTD as database for the BD.}

\end{center}
\label{fig:rest}
\end{figure*}

\begin{figure*}[ht]
\begin{center}
\includegraphics[width=0.15\linewidth]{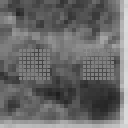}
\caption{\textbf{Right target is asked to be lighter.} LVI generated using RestoreNet as VTS and the DTD as database for the BD.}

\vspace{5cm}

\includegraphics[width=0.15\linewidth]{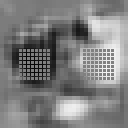}
\caption{\textbf{Right target is asked to be lighter.} LVI generated using RestoreNet as VTS and CIFAR10 as database for the BD.}

\vspace{5cm}

\includegraphics[width=0.15\linewidth]{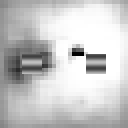}
\caption{\textbf{Right target is asked to be highlighted (higher contrast).} CrVI generated using RestoreNet as VTS and CIFAR10 as database for the BD.}

\end{center}
\label{fig:rest}
\end{figure*}

\begin{figure*}[ht]
\begin{center}

\includegraphics[width=0.15\linewidth]{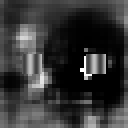}
\caption{\textbf{Right target is asked to be highlighted (higher contrast).} CrVI generated using RestoreNet as VTS and CIFAR10 as database for the BD.}

\vspace{5cm}

\includegraphics[width=0.15\linewidth]{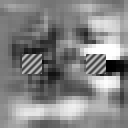}
\caption{\textbf{Right target is asked to be highlighted (higher contrast).} CrVI generated using RestoreNet as VTS and CIFAR10 as database for the BD.}

\end{center}
\label{fig:rest}
\end{figure*}


\begin{figure*}[ht]
\begin{center}
\includegraphics[width=0.15\linewidth]{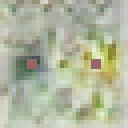}
\caption{\textbf{Right target is asked to be reddish.} CoVI generated using RestoreNet as VTS and the DTD as database for the BD.}

\vspace{5cm}

\includegraphics[width=0.15\linewidth]{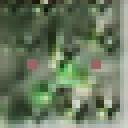}
\caption{\textbf{Right target is asked to be reddish.} CoVI generated using RestoreNet as VTS and the DTD as database for the BD.}

\vspace{5cm}

\includegraphics[width=0.15\linewidth]{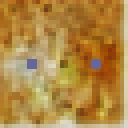}
\caption{\textbf{Right target is asked to be bluer.} CoVI generated using RestoreNet as VTS and the DTD as database for the BD.}

\end{center}
\label{fig:rest}
\end{figure*}



\hspace{30cm}
\newpage

\begin{figure*}[t]
\begin{center}
\includegraphics[width=0.15\linewidth]{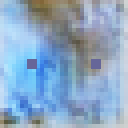}
\caption{\textbf{Right target is asked to be bluer.} CoVI generated using RestoreNet as VTS and the DTD as database for the BD.}

\vspace{5cm}

\includegraphics[width=0.15\linewidth]{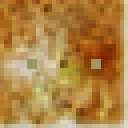}
\caption{\textbf{Right target is asked to be yellower.} CoVI generated using RestoreNet as VTS and the DTD as database for the BD.}

\vspace{5cm}

\includegraphics[width=0.15\linewidth]{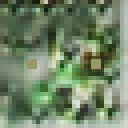}
\caption{\textbf{Right target is asked to be yellower.} CoVI generated using RestoreNet as VTS and the DTD as database for the BD.}

\end{center}
\label{fig:rest}
\end{figure*}



\vfill
\newpage

\begin{figure*}[t]
\begin{center}

\includegraphics[width=0.15\linewidth]{figs/dtd_assim.png}
\caption{\textbf{Right target is asked to be lighter.} LVI generated using RestoreNet as VTS and the DTD as database for the BD.}

\vspace{5cm}

\includegraphics[width=0.15\linewidth]{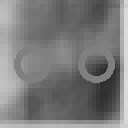}
\caption{\textbf{Right target is asked to be lighter.} LVI generated using RestoreNet as VTS and the DTD as database for the BD.}

\vspace{5cm}

\includegraphics[width=0.15\linewidth]{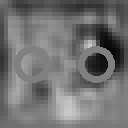}
\caption{\textbf{Right target is asked to be lighter.} LVI generated using RestoreNet as VTS and CIFAR10 as database for the BD.}

\end{center}
\label{fig:rest}
\end{figure*}





\end{appendices}

\end{document}